\documentclass[10pt,twocolumn,letterpaper]{article}

\usepackage{wacv}              %

\usepackage{amsmath}
\usepackage{amssymb}
\usepackage{booktabs}
\usepackage{enumitem}
\usepackage{graphicx}
\usepackage{multirow}
\usepackage{algorithm}
\usepackage{algpseudocode}

\usepackage{xcolor}
\usepackage{soul}

\usepackage[pagebackref,breaklinks,colorlinks]{hyperref}

\algnewcommand\algorithmicforeach{\textbf{for each}}
\algdef{S}[FOR]{ForEach}[1]{\algorithmicforeach\ #1\ \algorithmicdo}
 
\def\eg{\emph{e.g.}} 
\newcounter{linear}
\DeclareMathOperator*{\argmax}{arg\,max}

\usepackage[capitalize]{cleveref}
\crefname{section}{Sec.}{Secs.}
\Crefname{section}{Section}{Sections}
\Crefname{table}{Table}{Tables}
\crefname{table}{Tab.}{Tabs.}

\def\wacvPaperID{4} %
\def\confName{WACV}
\def\confYear{2025}

\begin{document}

\title{Addressing Out-of-Label Hazard Detection in Dashcam Videos: Insights from the COOOL Challenge}

\author{Anh-Kiet Duong\\
L3i Laboratory, La Rochelle University\\
17042 La Rochelle Cedex 1 - France\\
{\tt\small anh.duong@univ-lr.fr}
\and
Petra Gomez-Krämer\\
L3i Laboratory, La Rochelle University\\
17042 La Rochelle Cedex 1 - France\\
{\tt\small petra.gomez@univ-lr.fr}
}
\maketitle

\begin{abstract}
This paper presents a novel approach for hazard analysis in dashcam footage, addressing the detection of driver reactions to hazards, the identification of hazardous objects, and the generation of descriptive captions. We first introduce a method for detecting driver reactions through speed and sound anomaly detection, leveraging unsupervised learning techniques. For hazard detection, we employ a set of heuristic rules as weak classifiers, which are combined using an ensemble method. This ensemble approach is further refined with differential privacy to mitigate overconfidence, ensuring robustness despite the lack of labeled data. Lastly, we use state-of-the-art vision-language models for hazard captioning, generating descriptive labels for the detected hazards. Our method achieved the highest scores in the Challenge on Out-of-Label in Autonomous Driving, demonstrating its effectiveness across all three tasks. Source codes are publicly available at \url{https://github.com/ffyyytt/COOOL_2025}.
\end{abstract}

\section{Introduction}
\label{sec:intro}

Detecting hazards in dynamic environments is a critical task in computer vision, essential for real-time applications like autonomous driving, surveillance, and human-computer interaction. Accurate hazard detection requires identifying potential dangers, understanding their context, and describing them meaningfully. The COOOL challenge \cite{alshami2024coool} provides a unique benchmark by addressing the out-of-label problem, where annotations are limited to evaluation-only data. The challenge includes three tasks: the detection of driver state changes due to hazards, the identification of hazardous objects, and the generation of captions to describe these hazards. These tasks demand solutions that integrate temporal, spatial, and semantic reasoning, making it challenging and essential in computer vision.

Driver reaction detection and hazard detection can be framed as vision classification problems, where advanced deep learning models have demonstrated state-of-the-art performance. These models often require large labeled datasets for effective training \cite{liu2022swin, dosovitskiy2020image}. Additionally, Zero-Shot Learning (ZSL), which leverages deep learning models, can perform classification on labels that were not present in the training data. However, while ZSL offers flexibility, it often struggles with fine-grained task performance and lacks the domain-specific knowledge necessary for accurate detection \cite{guo2024fine}. The generation of captions has seen significant advancements in recent years, particularly with the rise of Large Language Models (LLMs), which have shown promise in bridging the gap between vision and language understanding \cite{li2023blip}. These developments have opened new possibilities for generating descriptive captions, although challenges remain in capturing key details throughout the entire video.

In this paper, we present our solution to the COOOL challenge \cite{alshami2024coool}, which achieved the first place. Our method addresses the out-of-label problem by leveraging advanced models and an optimized pipeline. By combining state-of-the-art vision and language models, we set new benchmarks on both the public and private leaderboards.

\section{Method}
In this section, we present our approach for hazard analysis in dashcam footage, addressing three tasks: detecting driver reactions (Section~\ref{sec:task1}), identifying hazardous objects (Section~\ref{sec:task2}), and generating hazard captions (Section~\ref{sec:task3}). We combine anomaly detection, heuristic rules, and vision-language models for robust and accurate detection, securing first place in the COOOL \cite{alshami2024coool, coolwacv25} challenge.

\label{sec:method}
\subsection{Driver reaction dectection}
\label{sec:task1}
To detect driver reactions to hazards, we propose two approaches: speed anomaly detection and sound anomaly detection. The first identifies sudden velocity changes, like abrupt braking or rapid acceleration, while the second captures anomalous sounds, such as shouting or emergency braking noises. Both methods rely on peak detection in unlabeled data \cite{shipmon2017time}, enabling effective anomaly detection without supervision. Together, they provide a robust solution for detecting driver reactions to hazards.

\subsubsection{Speed anomaly detection}
\label{sec:speed}
Driver reactions to hazards are often reflected in abrupt and irregular movements, such as sudden braking, rapid acceleration, or sharp changes in direction. These swift and unexpected maneuvers are critical indicators of hazardous events or challenging driving conditions. To effectively identify such reactions, we focus on detecting anomalies in the velocity profiles of objects and vehicles within the scene. However, directly analyzing object velocities within dashcam footage presents challenges due to the motion of the vehicle and the diverse movement patterns of surrounding objects, including those that are stationary, moving in the same direction, in the opposite direction, or other trajectories such as turning or crossing paths. Accounting for these varied behaviors is essential for robust and accurate anomaly detection in real-world driving environments.

To address these complexities, our method incorporates a two-stage approach. The steps of our approach are outlined in Algorithm~\ref{alg:speed_anomaly_detection}. First, for each object in the scene, we compute the centroid of its bounding box and measure its frame-to-frame displacement. This displacement is modeled linearly over time, with the slope representing the object's velocity. Second, recognizing the instability of using velocity alone, we compute the vehicle's acceleration. By treating the velocity-time relationship of the vehicle as a linear model, we extract the slope as its acceleration. By identifying peaks in the computed acceleration values (Line~\ref{alg:speed_anomaly_detection:return} of Algorithm~\ref{alg:speed_anomaly_detection}), we detect abrupt driver reactions indicative of hazardous events.

\begin{algorithm}[H]
\caption{Speed anomaly detection based on an object $o$}
\label{alg:speed_anomaly_detection}
\begin{algorithmic}[1]
\Require $F_o$: list of frame numbers where the object $o$ appears with annotated bounding boxes, $chunksize$: number of samples to compute the acceleration.
\State Compute the centroid of $o$ in the first frame: $C_{F_o\left[0\right]}$
\For{$f = 1$ \textbf{to} length $\left(F_o \right)$}
\State Compute the centroid of $o$: $C_{F_o\left[f\right]}$
\State Vehicle velocity to $o$ \footnotemark\setcounter{linear}{\value{footnote}}: $v_f \sim \frac{ \left\| C_{F_o\left[f\right]} - C_{F_o \left[0 \right]} \right\| }{f-F_o \left[0 \right]}$ 
\If{$f \ge chunksize$}
    \State Acceleration\footnotemark[\value{footnote}]: $a_f \sim \frac{ \left\| v_f - v_{F_o\left[f-chunksize \right]} \right\| }{chunksize} $ 
\EndIf
\If{ found peak in $a$ }
    \Return $f$ \label{alg:speed_anomaly_detection:return}
\EndIf
\EndFor
\end{algorithmic}
\end{algorithm}

\footnotetext{Instead of direct division, we use a linear regression model of the form $y = ax + b$, where $a$ represents the desired value (velocity/acceleration each case). This approach reduces noise caused by inaccuracies in bounding box annotations, providing more reliable results.}

In Algorithm~\ref{alg:speed_anomaly_detection}, the parameter \textit{chunksize} smooths the data by reducing noise from fluctuating bounding box coordinates, which can cause inaccurate velocity and acceleration estimates. A larger value of \textit{chunksize} improves the stability but may overlook brief, sudden events, while a smaller \textit{chunksize} increases the sensitivity but can lead to false-positive detections due to noise.

\subsubsection{Sound anomaly detection}
\label{sec:sound}
In addition to analyzing speed, audio signals provide a valuable dimension for detecting driver reactions to hazards. Sudden and unusual sounds, such as the driver shouting, emergency braking noises, or the honking of a horn, often accompany critical events on the road \cite{gatto2020audio}. These auditory cues are strong indicators of a driver's perception of danger or an imminent hazard. By leveraging the audio stream from dashcam recordings, anomalies in sound patterns can be identified and correlated with hazardous events, offering a complementary approach to speed-based detection methods. This multimodal analysis enhances the robustness of the system, particularly in scenarios where visual cues alone may not fully capture the driver's reactions.

To leverage sound data, we use an anomaly detection approach similar to speed analysis due to the lack of labels. Raw audio signals are preprocessed and normalized to reduce environmental noise (\eg, traffic sounds, background music) while preserving significant auditory cues like sudden loud noises or distinctive patterns. Peaks in the processed signal are identified to detect anomalies, such as shouting or emergency braking sounds, which indicate reactions to hazards. This unsupervised method enables the detection of critical auditory signals without relying on labeled data, enhancing the system's ability to capture diverse hazard scenarios.

\subsection{Hazard dectection}
\label{sec:task2}
Detecting hazards in dashcam footage presents significant challenges due to the absence of labeled data. However, several heuristic rules can be employed to create weak classifiers that do not require labeled datasets.
The following is a list of heuristic rules that we utilize as weak classifiers in our approach:

\begin{itemize}[topsep=1pt, parsep=1pt]
	\setlength{\itemsep}{1pt}
	\setlength{\parskip}{0pt}
	\setlength{\parsep}{0pt}
    \item Leverage pre-trained models on extensive datasets such as ImageNet \cite{deng2009imagenet, liu2022swin, dosovitskiy2020image} to classify objects and filter out those with labels that are unlikely to represent hazards, such as ``car'' or ``traffic light'', which frequently appear but are less critical as hazards.
    \item Evaluate the proximity of an object to the center of the video frame, as hazards are more likely to be near the center point of the dashcam.
    \item Analyze the frame-by-frame position of objects to determine whether their movement direction differs from the vehicle's trajectory, as objects with differing movement directions are more likely to pose hazards.
    \item Assess whether the object is actively participating in traffic. For this step, road lane detection algorithms can be utilized. However, due to time constraints, we approximate traffic regions by defining fixed zones within the video and considering objects within these zones as traffic participants.
    \item Examine the number of frames in which an object appears and the area of the object's bounding box. Large objects that persist across many frames are less likely to represent sudden or unexpected hazards.
    \item Correlate the appearance of objects with moments when the driver exhibits reactions (identified in Section~\ref{sec:task1}). Objects that appear close to these reaction points are more likely to be associated with hazards.
\end{itemize}

These individual rules are not entirely accurate in every situation, they offer a certain level of reliability. By combining these weak classifiers and leveraging diverse features, we construct a final model with improved performance and robustness. This ensemble approach mitigates the limitations of individual classifiers, enabling a more effective detection of hazards in complex driving scenarios.

To combine multiple weak classifiers, we employ a weighted ensemble approach \cite{dong2020survey}, where the weights are estimated based on the performance of each heuristic rule. However, due to the absence of labeled data, these weight estimations are inherently uncertain and may only perform well for specific videos while lacking generalizability across diverse scenarios. To address this issue and prevent overconfidence in any particular rule, we incorporate \textit{differential privacy} techniques. By perturbing the weights with noise drawn from a Gaussian distribution, controlled by a specified parameter $\epsilon$, we introduce robustness against overfitting to specific video contexts \cite{fraboni2024sifu, chen2023overconfidence}.

After perturbing the weights, we aggregate predictions from multiple noisy weight configurations using a voting ensemble \cite{dong2020survey}. In this setup, each object is assigned a score based on the number of votes it receives from these ensemble predictions. Objects with the highest number of votes are deemed the most likely hazards. This dual-layer ensemble approach enhances the model's robustness, leveraging the strengths of individual classifiers while mitigating the impact of uncertain weight estimates.

\subsection{Hazard captioning}
\label{sec:task3}
With the rapid advancements in large language models (LLMs) and vision-language models, generating descriptive captions for images and objects has become a powerful tool for understanding and interpreting visual data \cite{dinh2024trafficvlm}. In this task, we leverage these developments to generate captions for identified hazards in dashcam footage. The objective is to provide meaningful and context-aware descriptions that help characterize the detected hazards, offering insights into their nature and potential risks.

To achieve this, we utilize state-of-the-art image captioning models such as BLIPv2 \cite{li2023blip}, BLIP \cite{li2022blip}, and CLIP \cite{radford2021learning}, which are designed to generate captions based on the visual features of an image. These models are adept at recognizing a wide variety of objects and describing their attributes, making them ideal for our task. The process of generating captions is detailed in Algorithm~\ref{alg:hazard_caption}, which illustrates how captions are assigned to each hazard based on bounding box areas across frames.

\begin{algorithm}[H]
\caption{Hazard captioning for object $o$}
\label{alg:hazard_caption}
\begin{algorithmic}[1]
\Require $F_o$: list of frame indices where the object $o$ appears, $bbox_o[i]$: cropped bounding box image of the object $o$ at the $i^{th}$ frame, $M$: list of captioning models, $R$: hash table initialized to $0$.

\For{$f = 1$ \textbf{to} $\text{length}(F_o)$}
    \ForEach{captioning model $m$ in $M$}
        \State Generate caption: $\texttt{text} \gets m\left(bbox_o\left[F_o[f]\right]\right)$
        \State Compute area: $A \gets \text{width}(bbox_o[F_o[f]]) \times \text{height}(bbox_o[F_o[f]])$
        \State Update score in hash table: $R[\texttt{text}] \gets R[\texttt{text}] + A$
    \EndFor
\EndFor

\Return $\argmax_{\texttt{text} \in R} \, R[\texttt{text}]$
\end{algorithmic}
\end{algorithm}

In Algorithm~\ref{alg:hazard_caption}, for each frame where the object appears, its bounding box is cropped, and a caption is generated using each captioning model $m$. The area of the bounding box is calculated and used as a weight to update a hash table that accumulates scores for each unique caption. At the end of the process, the caption with the highest score in the hash table is selected as the final description for the hazard. This scoring mechanism prioritizes captions associated with larger or more prominently visible bounding boxes, ensuring that the most relevant description is chosen.

The evaluation metric for this task only considers the first 35 characters of a caption and searches the ground-truth annotations for their presence. To improve we refine the scoring process in Algorithm~\ref{alg:hazard_caption} to operate at the word level instead of the entire text. Instead of assigning scores to entire captions, we distribute the bounding box area among each individual words in the text. Consequently, the hash table $R$ now stores word-score mappings rather than text-score mappings. Nouns and meaningful words are given extra by doubling their scores, while common stop words (e.g., "a", "an", "the") are reduced \cite{silva2003importance}. Then, we double the score for objects that do not belong on the street (e.g., animals, trees). After scoring, we construct the final caption by selecting the highest-scoring words until the 35-character limit is met. This refinement ensures the selected captions are concise and aligned with the evaluation requirements. However, it may reduce the readability and interpretability of the generated captions for human users. As a result, we employ this refinement exclusively for metric optimization and do not integrate it into Algorithm~\ref{alg:hazard_caption}. Instead, the algorithm retains its original design for broader applicability and clearer caption generation.

\section{Experiments}
In this section, we present the details of the COOOL dataset used for the challenge and summarize the experimental setup. We also provide the results of various methods applied to the three tasks, along with their performance on both the private and public leaderboards.
\label{sec:results}
\subsection{Challenge description}

\begin{figure}[h]
    \centering
    \includegraphics[width=0.9\linewidth]{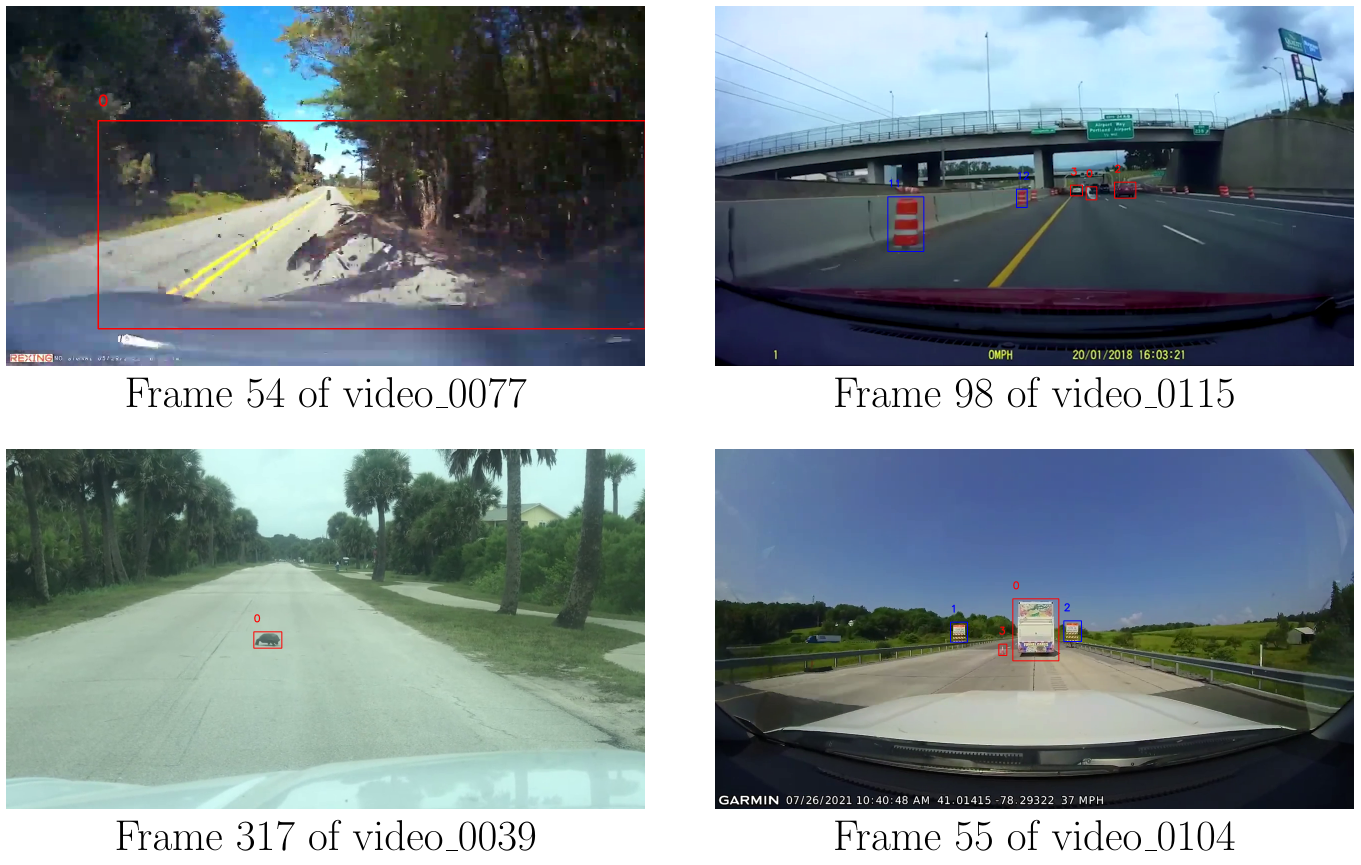}
    \caption{Sample frames from some videos of the COOOL dataset \cite{alshami2024coool}. The \textcolor{red}{red} bounding box denotes the \textit{challenge\_object}, while the \textcolor{blue}{blue} bounding box represents the \textit{traffic\_scene} as labeled in the \texttt{annotations\_public} file. The number of each bounding box corresponds to the tracking ID of the respective object.}
    \label{fig:dataset}
\end{figure}

The COOOL dataset \cite{alshami2024coool} consists of 200 video clips with annotations in the \texttt{annotations\_public} file, which includes bounding boxes for \textit{challenge\_objects} and \textit{traffic\_scenes}, each with a unique tracking ID. Sample frames from the dataset are shown in Figure~\ref{fig:dataset}, where the red and blue bounding boxes correspond to the \textit{challenge\_object} and \textit{traffic\_scene} annotations, respectively. Unlike conventional datasets with fully labeled data, COOOL is an evaluation-only benchmark with sparse annotations, requiring participants to infer information without comprehensive labels. This setup increases task complexity and requires methods that generalize effectively. Participants are evaluated on three key challenges using the following metrics:

\begin{itemize}
    \item Score for driver reaction detection:
    \begin{equation*}
        \tiny
        \frac{correct\_state\_change\_prediction}{total\_frames}
    \end{equation*}
    \item Score of correctly identified bounding box(es) containing hazards (average across all frames):
    \begin{equation*}
        \tiny
         \frac{correct\_predicted\_hazards}{\max \left( known\_hazards, len \left(predicted\_hazards \right) \right)}
    \end{equation*}
    \item Score for description (average across all frames):
    \begin{equation*}
        \tiny
        \frac{correct\_predicted\_caption}{\max \left( known\_caption, len \left(predicted\_caption \right) \right)}
    \end{equation*}
\end{itemize}

\subsection{Results}
The results of the experiments are summarized in Table~\ref{tab:results}. The table shows the performance of various methods on the three tasks on both the private and public leaderboards. The public leaderboard uses 8\% of the data, while the private leaderboard includes the remaining 92\%. Notably, the method combining all tasks achieved the highest scores on both leaderboards, securing the first place overall in the challenge. As this is a new competition, there are limited existing methods available for direct comparison.

\begin{table}[h]
\caption{Performance results for various methods on the COOOL challenge tasks. Where "False" or "-1" represent fixed values assigned to the predictions for that task.}
\label{tab:results}
\resizebox{\linewidth}{!}{\begin{tabular}{ccc|cc}
\hline
\multicolumn{3}{c|}{Method}                          & \multicolumn{2}{c}{Score} \\ \hline
Task 1        & Task 2          & Task 3           & Private     & Public      \\ \hline
speed (\ref{sec:speed})         & -1              & -1                & 0.25340     & 0.27579     \\
sound (\ref{sec:sound}) + speed & -1              & -1                & 0.27534     & 0.28458     \\ \hline
False         & proposed method (\ref{sec:task2}) & -1                & 0.29901     & 0.35704     \\ \hline
sound + speed & proposed method & blip2-opt-6.7b \cite{li2023blip, zhang2022opt}    & 0.51694     & 0.76830     \\
sound + speed & proposed method & blip2-flan-t5-xxl \cite{li2023blip, chung2024scaling} & 0.51663     & 0.69252     \\
sound + speed & proposed method & blip \cite{li2022blip}             & 0.49240     & 0.61384     \\
sound + speed & proposed method & vit-gpt2 \cite{kumar2022imagecaptioning}         & 0.47598     & 0.59966     \\ \hline
Baseline      & Baseline        & Baseline          & 0.25560     & 0.25681     \\
sound + speed & proposed method & all model         & 0.57261     & 0.78453    
\end{tabular}}
\end{table}

The baseline scores in Table~\ref{tab:results} correspond to the default method provided by the challenge organizers. In this baseline approach, the first task is based on a model that detects if the velocity is negative compared to other objects, the second task identifies the object closest to the center, and the third task uses the CLIP \cite{radford2021learning} model for caption generation.

\section{Conclusion} 
\label{sec:conclusion}
We have proposed an effective hazard analysis framework for dashcam footage, addressing three key tasks: detecting driver reactions to hazards, identifying hazardous objects, and generating descriptive captions for these hazards. By leveraging an ensemble of multiple methods for each task, our approach ensures robust and stable performance in hazard analysis. This method, tested on the COOOL dataset, demonstrates the effectiveness of combining various strategies to tackle hazard analysis in real-world driving scenarios.

{\small
\bibliographystyle{ieee_fullname}
\bibliography{references}
}

\end{document}